\documentclass[twoside]{article} 

\usepackage{authblk}

\usepackage{style_arXiv}

\usepackage{natbib}
\usepackage[utf8]{inputenc}
\usepackage{amssymb}
\usepackage{amsmath,amsthm}
\usepackage{mathtools}
\usepackage{amsfonts}
\usepackage{stmaryrd}
\usepackage{subfigure}
\usepackage{pgf}
\usepackage{algorithm}
\usepackage{algorithmic}

\usepackage{enumitem}

\newcommand\myexp{\mathop{{}\mathbb{E}}}

\DeclareMathOperator{\tr}{tr}
\DeclareMathOperator*{\argmin}{arg\,min}

\makeatletter
\newtheorem*{rep@theorem}{\rep@title}
\newcommand{\newreptheorem}[2]{%
\newenvironment{rep#1}[1]{%
 \def\rep@title{#2 \ref{##1}}%
 \begin{rep@theorem}}%
 {\end{rep@theorem}}}
\makeatother

\newcommand{\costcol}{\mathcal Q_{\mathit{CL}}} % Collaborative cost
\newcommand{\costmp}{\mathcal Q_{\mathit{MP}}} % Model propagation cost
\newcommand{\R}{\mathbb{R}}             % Reals
\newcommand{\loc}{\mathit{sol}}         % (superscript) local abrev.
\newcommand{\loss}{\ell}                   % Loss function (parametrized)
\newcommand{\Loss}{\mathcal{L}}                   % Global Loss function (parametrized)
\newcommand{\intset}[1]{\llbracket #1\rrbracket}
\newcommand{\T}{^{\top}}
\newcommand{\ttheta}{\widetilde{\Theta}}
\newcommand{\Nei}[1]{\mathcal{N}_{#1}}
\newcommand{\balpha}{\bar{\alpha}}

\newtheorem{theorem}{Theorem} 

\newtheorem{proposition}{Proposition}
\newreptheorem{theorem}{Theorem} 
\newreptheorem{lemma}{Lemma}
\newreptheorem{proposition}{Proposition}

\usepackage[pdfborder={0 0 0}]{hyperref}

\usepackage{xifthen}
\usepackage{xspace}

\def\arxiv{0}

\newcommand{\suppapp}[2]{%
  \ifthenelse{\arxiv=0}%
    {#1\xspace}% if #1 is empty
    {#2\xspace}% if #1 is not empty
}

\def\arxiv{1}

\begin{document}

\title{Decentralized Collaborative Learning of\\ Personalized Models over Networks}
\author[1]{Paul Vanhaesebrouck\thanks{\texttt{first.last@gmail.com}}}
\author[1]{Aurélien Bellet\thanks{\texttt{first.last@inria.fr}}}
\author[2]{Marc Tommasi$^\dagger$}
\affil[1]{INRIA}
\affil[2]{Université de Lille}
\date{}
\maketitle

% !TEX root = main.tex
\begin{abstract}

We consider a set of learning agents in a collaborative peer-to-peer
network, where each agent learns a \emph{personalized model}
according to its own learning objective. The question addressed in
this paper is: how can agents improve upon their locally trained
model by communicating with other agents that have similar
objectives? We introduce and analyze two asynchronous gossip
algorithms running in a fully decentralized manner. Our first
approach, inspired from label propagation, aims to smooth pre-trained local models over the network while accounting for the confidence that each agent has in its initial model. In our second
approach, agents jointly learn and propagate their
model by making iterative updates based on both their local dataset
and the behavior of their neighbors. To optimize this challenging objective, our decentralized algorithm is based on ADMM.

  % We consider a set of learning agents, each with their own personal data, which participate in a collaborative peer-to-peer network.
  % Unlike existing work focusing on problems where agents seek to agree on a \emph{consensus model}, we study the setting where each agent learns a \emph{personalized model} according to its own learning objective. The question addressed in this paper is: how can agents improve upon their locally trained model by communicating with other agents that have similar objectives?
  % Our first approach propagates pre-trained local models across the network, taking into account the confidence that each agent has in its own model.
  % In our second approach, agents jointly learn and propagate their model by making iterative updates based on both their local dataset and the behavior of their neighbors in the network.
  % To allow an efficient deployment of both approaches in real networks, we introduce and analyze asynchronous gossip algorithms running in a fully decentralized manner.
  % Their performance is evaluated through a series of experiments on a set of synthetic tasks.
\end{abstract}

%%% Local Variables:
%%% mode: latex
%%% ispell-local-dictionary: "english"
%%% TeX-master: "main"
%%% End:

% !TEX root = main.tex

\section{Introduction}
\label{sec:introduction}

Increasing amounts of data are being produced by interconnected devices such as mobile phones, connected objects, sensors, \emph{etc}. For instance, history logs are generated when a smartphone user browses the web, gives product ratings and executes various applications. The currently dominant approach to extract useful information from such data is to collect all users' personal data on a server (or a tightly coupled system hosted in a data center) and apply centralized machine learning and data mining techniques.
However, this centralization poses a number of issues, such as the need for users to ``surrender'' their personal data to the service provider without much control on how the data will be used, while incurring potentially high bandwidth and device battery costs. Even when the learning algorithm can be distributed in a way that keeps data on users' devices, a central entity is often still required for aggregation and coordination \citep[see e.g.,][]{McMahan2016a}.

In this paper, we envision an alternative setting where many users (agents) with local datasets collaborate to learn models by engaging in a fully decentralized peer-to-peer network. Unlike existing work focusing on problems where agents seek to agree on a global \emph{consensus model} \citep[see e.g.,][]{Nedic2009a,Wei2012a,Duchi2012a}, we study the case where each agent learns a \emph{personalized model} according to its own learning objective. We assume that the network graph is given and reflects a notion of similarity between agents (two agents are neighbors in the network if they have a similar learning objective), but each agent is only aware of its direct neighbors.
An agent can then learn a model from its (typically scarce) personal data but also from interactions with its neighborhood.
As a motivating example, consider a decentralized recommender system \citep{Boutet2013a,Boutet2014a} in which each user rates a small number of movies on a smartphone application and expects personalized recommendations of new movies. In order to train a reliable recommender for each user, one should rely on the limited user's data but also on information brought by users with similar taste/profile. The peer-to-peer communication graph could be established when some users go the same movie theater or attend the same cultural event, and some similarity weights between users could be computed based on historical data (e.g., counting how many times people have met in such locations). 

Our contributions are as follows. After formalizing the problem of interest, we propose two asynchronous and fully decentralized algorithms for collaborative learning of personalized models. They belong to the family of gossip algorithms \citep{Shah2009a,Dimakis2010a}: agents only communicate with a single neighbor at a time, which makes our algorithms suitable for deployment in large peer-to-peer real networks.
Our first approach, called \emph{model propagation}, is inspired by the graph-based label propagation technique of \citet{zhou2004learning}. In a first phase, each agent learns a model based on its local data only, without communicating with others. In a second phase, the model parameters are regularized so as to be smooth over the network graph. We introduce some confidence values to account for potential discrepancies in the agents' training set sizes, and derive a novel asynchronous gossip algorithm which is simple and efficient. We prove that this algorithm converges to the optimal solution of the problem.
Our second approach, called \emph{collaborative learning}, is more flexible as it interweaves learning and propagation in a single process. Specifically, it optimizes a trade-off between the smoothness of the model parameters over the network on the one hand, and the models' accuracy on the local datasets on the other hand. For this formulation, we propose an asynchronous gossip algorithm based on a decentralized version of Alternating Direction Method of Multipliers (ADMM) \citep{boyd2011distributed}.
Finally, we evaluate the performance of our methods on two synthetic collaborative tasks: mean estimation and linear classification. Our experiments show the superiority of the proposed approaches over baseline strategies, and confirm the efficiency of our decentralized algorithms.

The rest of the paper is organized as follows. Section~\ref{sec:setting} formally describes the problem of interest and discusses some related work. Our model propagation approach is introduced in Section~\ref{sec:modelpropagation}, along with our decentralized algorithm. Section~\ref{sec:colearning} describes our collaborative learning approach, and derives an equivalent formulation which is amenable to optimization using decentralized ADMM. Finally, Section~\ref{sec:experiments} shows our numerical results, and we conclude in Section~\ref{sec:conclu}.

%%% Local Variables:
%%% mode: latex
%%% ispell-local-dictionary: "english"
%%% TeX-master: "main"
%%% End:

% !TEX root = main.tex

\section{Preliminaries}
\label{sec:setting}

\subsection{Notations and Problem Setting}

% \begin{figure}[t]
%   \centering
%   \includegraphics[width=.7\columnwidth]{./images/fig.pdf}
%   \caption{Illustration of our problem setting. Agents are organized in a network with edge weights indicating the similarity between agents' target model. Each agent has its own dataset of potentially different size.}
%   \label{fig:setting}
% \end{figure}

We consider a set of $n$ agents $V=\intset{n}$ where  $\intset{n}:=\{1,\dots,n\}$. Given a convex loss function $\loss:\mathbb{R}^p\times\mathcal{X}\times\mathcal{Y}$, the goal of agent $i$ is to learn a model $\theta_i\in\mathbb{R}^p$ whose expected loss $\myexp_{(x_i,y_i)\sim\mu_i} \loss(\theta_i; x_i,y_i)$ is small with respect to an unknown and fixed  distribution $\mu_i$  over $\mathcal{X}\times\mathcal{Y}$.
Each agent $i$ has access to a set of $m_i\geq 0$ i.i.d. training examples $\mathcal{S}_i=\{(x_i^j,y_i^j)\}_{j=1}^{m_i}$ drawn  from  $\mu_i$. We allow the training set size to vary widely across agents (some may even have no data at all). This is important in practice as some agents may be more ``active'' than others, may have recently joined the service, \emph{etc}.  

In isolation, an agent $i$ can learn a \emph{``solitary'' model} $\theta_i^{\loc}$ by minimizing the loss over its local dataset $\mathcal{S}_i$: 
\begin{equation}
\label{eq:local_model}
\theta_i^{\loc} \in\argmin_{\theta\in\mathbb{R}^p} \Loss_i(\theta) = \sum_{j=1}^{m_i} \loss(\theta; x_i^j,y_i^j).
\end{equation}
The goal for the agents is to improve upon their solitary model by leveraging information from other users in the network.
% which represent the underlying relationship between the agents' objectives.
Formally, we consider a weighted connected graph $G=(V,E)$ over the set $V$ of agents, where $E\subseteq V\times V$ is the set of undirected edges.
We denote by $W \in \mathbb{R}^{n \times n}$ the symmetric nonnegative weight matrix associated with $G$, where $W_{ij}$ gives the weight of edge $(i,j)\in E$ and by convention, $W_{ij}=0$ if $(i,j)\notin E$ or $i=j$.
We assume that the weights represent the underlying similarity between the agents' objectives: $W_{ij}$ should tend to be large (resp. small) when the objectives of agents $i$ and $j$ are similar (resp. dissimilar).
While we assume in this paper that the weights are given, in practical scenarios one could for instance use some auxiliary information such as users' profiles (when available) and/or prediction disagreement to estimate the weights.
For notational convenience, we define the diagonal matrix $D\in\mathbb{R}^{n\times n}$ where $D_{ii} = \sum_{j=1}^n W_{ij}$.
We will also denote by $\mathcal{N}_i = \{j\neq i : W_{ij}>0\}$ the set of neighbors of agent $i$. We assume that the agents only have a local view of the network: they know their neighbors and the associated weights, but not the global topology or how many agents participate in the network.

%The problem setting is illustrated in Figure~\ref{fig:setting}. 
Our goal is to propose decentralized algorithms for agents to collaboratively improve upon their solitary model by leveraging information from their neighbors. 

\subsection{Related Work}

Several peer-to-peer algorithms have been developed for decentralized averaging \citep{Kempe2003a,boyd2006randomized,Colin2015a} and optimization \citep{Nedic2009a,Ram2010a,Duchi2012a,Wei2012a,Wei2013a,Iutzeler2013a,Colin2016a}. These approaches solve a \emph{consensus problem} of the form:
\begin{equation}
\label{eq:consensus}
\min_{\theta\in\mathbb{R}^p} \sum_{i=1}^n \Loss_i(\theta),
\end{equation}
resulting in a global solution common to all agents (e.g., a classifier minimizing the prediction error over the union of all datasets). This is unsuitable for our setting, where all agents have personalized objectives.

Our problem is reminiscent of Multi-Task Learning (MTL) \citep{Caruana1997a}, where one jointly learns models for related tasks. Yet, there are several differences with our setting. In MTL, the number of tasks is often small, training sets are well-balanced across tasks, and all tasks are usually assumed to be positively related (a popular assumption is that all models share a common subspace). Lastly, the algorithms are centralized, aside from the distributed MTL of \citet{Wang2016a} which is synchronous and relies on a central server.

%%% Local Variables:
%%% mode: latex
%%% ispell-local-dictionary: "english"
%%% TeX-master: "main"
%%% End:

% !TEX root = main.tex

\section{Model Propagation}
\label{sec:modelpropagation}

In this section, we present our model propagation approach. % which performs a smoothing of the initial local models $\theta_1^{\loc},\dots,\theta_n^{\loc}$ over the network.
We first introduce a global optimization problem, and then propose and analyze an asynchronous gossip algorithm to solve it.

\subsection{Problem Formulation}
\label{ssec:mp_formulation}

In this formulation, we assume that each agent~$i$ has learned a solitary model $\theta_i^{\loc}$ by minimizing its local loss, as in \eqref{eq:local_model}. This can be done without any communication between agents.
Our goal here consists in adapting these models by making them smoother over the network graph. In order account for the fact that the solitary models were learned on training sets of different sizes, we will use $c_i \in(0,1]$ to denote the confidence we put in the model $\theta_i^{\loc}$ of user $i\in\{1,\dots,n\}$. The $c_i$'s should be proportional to the number of training points $m_i$ --- one may for instance set $c_i=\frac{m_i}{\max_{j} m_j}$ (plus some small constant in the case where $m_i=0$).

Denoting $\Theta = [\theta_1;\dots;\theta_n]\in\mathbb{R}^{n\times p}$, the objective function we aim to minimize is as follows:
\begin{multline}
    \costmp(\Theta) = \frac{1}{2} \bigg( \sum_{i<j}^n W_{ij} {\lVert \theta_i - \theta_j \rVert}^2 + \\ \mu \sum_{i=1}^n D_{ii} c_i {\lVert \theta_i - \theta_i^{\loc} \rVert}^2 \bigg),
    \label{eq:Qw}
\end{multline}
where $\mu > 0$ is a trade-off parameter and $\lVert \cdot \rVert$ denotes the Euclidean norm.
The first term in the right hand side of \eqref{eq:Qw} is a classic %Laplacian 
quadratic form used to smooth the models within neighborhoods: the distance between the new models of agents $i$ and $j$ is encouraged to be small when the weight $W_{ij}$ is large. The second term prevents models with large confidence from diverging too much from their original values so that they can propagate useful information to their neighborhood. On the other hand, models with low confidence are allowed large deviations: in the extreme case where agent $i$ has very little or even no data (i.e., $c_i$ is negligible), its model is fully determined by the neighboring models. The presence of $D_{ii}$ in the second term is simply for normalization.
We have the following result (the proof is in \suppapp{the supplementary material}{\ref{app:prop1}}).

\begin{proposition}[Closed-form solution]
\label{prop:MPclosedform}
Let $P = D^{-1}W$ be the stochastic similarity matrix associated with the graph $G$ and $\Theta^{\loc} = [\theta^{\loc}_1;\dots;\theta^{\loc}_n]\in\mathbb{R}^{n\times p}$. The solution ${\Theta^\star = \argmin_{\Theta \in \mathbb{R}^{n \times p}} \costmp(\Theta)}$ is given by
\begin{equation}
     \Theta^\star = \balpha {(I - \balpha (I-C) - \alpha P)}^{-1} C\Theta^{\loc},
    \label{eq:mp_closedform}
\end{equation}
with $\alpha\in(0,1)$ such that $\mu = (1 - \alpha) / \alpha$, and $\balpha=1-\alpha$.
\end{proposition}

Our formulation is a generalization of the semi-supervised label propagation technique of \citep{zhou2004learning}, which can be recovered by setting $C=I$ (same confidence for all nodes). Note that it is \emph{strictly} more general: we can see from \eqref{eq:mp_closedform} that unless the confidence values are equal for all agents, the confidence information cannot be incorporated by using different solitary models $\Theta^{\loc}$ or by considering a different graph (because $\frac{\balpha}{\alpha}(I-C)-P$ is not stochastic). The asynchronous gossip algorithm we present below thus applies to label propagation for which, to the best of our knowledge, no such algorithm was previously known.

Computing the closed form solution \eqref{eq:mp_closedform} requires the knowledge of the global network and of all solitary models, which are unknown to the agents.
% Next, we propose a decentralized and asynchronous iterative algorithm which converges to the same solution.
Our starting point for the derivation of an asynchronous gossip algorithm is the following iterative form: for any $t\geq 0$,
\begin{equation}
    \Theta(t+1) = {(\alpha I + \balpha  C)}^{-1} \left(\alpha P \Theta(t) + \balpha C\Theta^{\loc}\right),
    \label{eq:modelit}
\end{equation}
The sequence ${(\Theta(t))}_{t \in \mathbb{N}}$ can be shown to converge to \eqref{eq:mp_closedform} regardless of the choice of initial value $\Theta(0)$, see \suppapp{supplementary material}{\ref{app:iter}} for details.
An interesting observation about this recursion is that it can be decomposed into agent-centric updates which only involve neighborhoods. Indeed, for any agent $i$ and any $t\geq 0$:
\begin{equation*}
    \theta_i(t+1) = \frac{1}{\alpha + \balpha c_i} \Big( \alpha \sum_{j \in \mathcal{N}_i} \frac{W_{ij}}{D_{ii}} \theta_j(t) + \balpha c_i \theta^{\loc}_i \Big).
\end{equation*}
%Remark that $\theta_i(t+1)$ does not depend on $\theta(t)_i$, this comes from the fact that updating $\theta_i$ only would result in finding the minimum of~\eqref{eq:Qw} with all other coordinates $\theta_j$ for $j \neq i$ fixed.
The iteration \eqref{eq:modelit} can thus be understood as a decentralized but synchronous process where, at each step, every agent communicates with all its neighbors to collect their current model parameters and uses this information to update its model.
Assuming that the agents do have access to a global clock to synchronize the updates (which is unrealistic in many practical scenarios), synchronization incurs large delays since all agents must finish the update at step~$t$ before anyone starts step~$t+1$. The fact that agents must contact all their neighbors at each iteration further hinders the efficiency of the algorithm.
% This requires that the agents have access to a global clock to synchronize the updates, but synchronization is typically a very expensive operation in decentralized networks (if possible at all). Furthermore, all agents must wait for all nodes to finish the update at step~$t$ before starting step~$t+1$.
To avoid these limitations, we propose below an asynchronous gossip algorithm.

\subsection{Asynchronous Gossip Algorithm}
\label{ssec:asyncmp}

In the asynchronous setting, each agent has a \emph{local} clock ticking at the times of a rate 1 Poisson process, and wakes up when it ticks. As local clocks are i.i.d., it is equivalent to activating a single node uniformly at random at each time step \citep{boyd2006randomized}.\footnote{Our analysis straightforwardly extends to the case where agents have clocks ticking at different rates.}

The idea behind our algorithm is the following. At any time $t\geq 0$, each agent $i$ will maintain a (possibly outdated) knowledge of its neighbors' models.  For mathematical convenience, we will consider a matrix $\ttheta_i(t)\in\R^{n\times p}$ where its $i$-th line ${\ttheta_i^i(t)\in\R^p}$ is agent $i$'s model at time $t$, and for $j\neq i$, its $j$-th line $\ttheta_i^j(t)\in\mathbb{R}^p$ is agent $i$'s \emph{last knowledge} of the model of agent~$j$. For any $j\notin\mathcal{N}_i \cup \{i\}$ and any $t\geq 0$, we will maintain $\ttheta_i^j(t)=0$.  Let $\ttheta=[\ttheta_1^\top , \dots , \ttheta_n^\top]^\top\in\R^{n^2\times p}$ be the horizontal stacking of all the $\ttheta_i$'s. 

If agent $i$ wakes up at time step~$t$, two consecutive actions are performed:
\begin{itemize}
\item \emph{communication step}: agent $i$ selects a random neighbor $j\in\mathcal{N}_i$ with prob. $\pi_i^j$ and both agents update their knowledge of each other's model:
$$\ttheta_i^j(t+1) = \ttheta^j_j(t) \text{ and } \ttheta_j^i(t+1) = \ttheta^i_i(t), $$
\item \emph{update step}: agents $i$ and $j$ update their own models based on current knowledge. For $l\in\{i,j\}$:
\begin{multline}
\label{eq:mp_update}
    \ttheta_l^l(t+1) = {(\alpha + \balpha c_l)}^{-1} \\ \Big( \alpha \sum_{k \in \mathcal{N}_l} \frac{W_{lk}}{D_{ll}} \ttheta_l^k(t+1) + \balpha c_l \theta^{\loc}_l \Big).
\end{multline}
\end{itemize}
All other variables in the network remain unchanged. In the communication step above, $\pi_i^j$ corresponds to the probability that agent $i$ selects agent $j$. For any $i\in\intset{n}$, we have $\pi_i\in{[0,1]}^n$ such that $\sum_{j=1}^n\pi_i^j=1$ and $\pi_i^j>0$ if and only if $j \in \Nei{i}$.  % For any agent $i$ and any time step $t$, we denote the probabilities for communicating with its neighbors by $\pi_i\in\mathbb{R}^n$, where $\pi_i^j$ corresponds to the probability that agent $i$ communicates with agent $j$.

Our algorithm belongs to the family of \emph{gossip algorithms} as each agent communicates with at most one neighbor at a time. Gossip algorithms are known to be very effective for decentralized computation in peer-to-peer networks \cite[see][]{Dimakis2010a,Shah2009a}. Thanks to its asynchronous updates, our algorithm has the potential to be much faster than a synchronous version when executed in a large peer-to-peer network.

The main result of this section shows %the convergence of our decentralized model propagation algorithm.
that our algorithm converges to a state where all nodes have their optimal model (and those of their neighbors).
% The first important step is to rewrite the algorithm as an equivalent  random iterative process over $\ttheta\in\mathbb{R}^{n^2\times p}$: 
% \begin{equation}
% \label{eq:globalalg}
% \text{for any $t\geq 0$}\quad \ttheta(t+1) = A(t)\ttheta(t) + b(t).
% \end{equation}

\begin{theorem}[Convergence]
\label{thm:asyncMP}
Let $\ttheta(0)\in\mathbb{R}^{n^2\times p}$ be some arbitrary initial value and $(\ttheta(t))_{t\in\mathbb{N}}$ be the sequence generated by our algorithm. Let $ \Theta^\star = \argmin_{\Theta \in \mathbb{R}^{n \times p}} \costmp(\Theta)$ be the optimal solution to model propagation. For any $i\in\intset{n}$, we have:
$$\lim\limits_{t \rightarrow \infty} \mathbb{E} \left[ \ttheta_i^j(t) \right] =  \Theta^\star_j \text{ for } j \in \mathcal{N}_i \cup \{i\}.$$
\end{theorem}
\begin{proof}[Sketch of proof]
The first step of the proof is to rewrite the algorithm as an equivalent random iterative process over $\ttheta\in\mathbb{R}^{n^2\times p}$ of the form:
\begin{equation*}
%\label{eq:globalalg}
\ttheta(t+1) = A(t)\ttheta(t) + b(t),
\end{equation*}
for any $t\geq 0$. Then, we show that the spectral radius of $\mathbb{E}[A(t)]$ is smaller than 1, which allows us to exhibit the convergence to the desired quantity. The proof can be found in \suppapp{the supplementary material}{\ref{app:async_mp}}.
\end{proof}

% The proof can be found in the supplementary material. 
%Theorem~\ref{thm:asyncMP} shows that our asynchronous algorithm converges to a state where all nodes have their optimal model (and those of their neighbors).
%In our simulations (Section~\ref{sec:experiments}), we will show that this algorithm is competitive with the synchronous version in terms of the number of communications needed to converge. Thanks to its asynchronous updates, the proposed algorithm thus has the potential to be much faster than the synchronous version when used in a real network.
%%% Local Variables:
%%% mode: latex
%%% ispell-local-dictionary: "english"
%%% TeX-master: "main"
%%% End:

% !TEX root = main.tex

\section{Collaborative Learning}
\label{sec:colearning}

In the approach presented in the previous section, models are learned locally by each agent and then propagated through the graph. % in a peer-to-peer fashion.
In this section, we allow the agents to simultaneously learn their model and propagate it through the network. In other words, agents iteratively update their models based on both their local dataset and the behavior of their neighbors. While in general this is computationally more costly than merely propagating pre-trained models, we can expect significant improvements in terms of accuracy.

As in the case of model propagation, we first introduce the global objective function and then propose an asynchronous gossip algorithm, which is based on the general paradigm of ADMM \citep{boyd2011distributed}. 

\subsection{Problem Formulation}

In contrast to model propagation, the objective function to minimize here takes into account the loss of each personal model on the local dataset, rather than simply the distance to the solitary model:
\begin{equation}
    \label{eq:Qcl}
    \costcol(\Theta) = \sum_{i<j}^n W_{ij} {\lVert \theta_i - \theta_j \rVert}^2 + \mu \sum_{i=1}^n D_{ii} \Loss_i(\theta_i),
\end{equation}
where $\mu>0$ is a trade-off parameter. The associated optimization problem is $\Theta^\star = \argmin_{\Theta \in \mathbb{R}^{n \times p}} \costcol(\Theta)$.

The first term in the right hand side of \eqref{eq:Qcl} is the same as in the model propagation objective \eqref{eq:Qw} and tends to favor models that are smooth on the graph. However, while in model propagation enforcing smoothness on the models may potentially translate into a significant decrease of accuracy on the local datasets (even for relatively small changes in parameter values with respect to the solitary models), here the second term prevents this. It allows more flexibility in settings where very different parameter values define models which actually give very similar predictions. Note that the confidence is built in the second term as $\Loss_i$ is a sum over the local dataset of agent $i$.

In general, there is no closed-form expression for $\Theta^\star$, but we can solve the problem with a decentralized iterative algorithm, as shown in the rest of this section.

\subsection{Asynchronous Gossip Algorithm}
\label{ssec:async_cl}

We propose an asynchronous decentralized algorithm for minimizing \eqref{eq:Qcl} based on the Alternative Direction Method of Multipliers (ADMM). This general method is a popular way to solve consensus problems of the form \eqref{eq:consensus} in the distributed and decentralized settings  \citep[see e.g.,][]{boyd2011distributed,Wei2012a,Wei2013a,Iutzeler2013a}.
In our setting, we do not seek a consensus in the classic sense of \eqref{eq:consensus} since our goal is to learn a personalized model for each agent. However, we show below that we can reformulate \eqref{eq:Qcl} as an equivalent \emph{partial} consensus problem which is amenable to decentralized optimization with ADMM.
%on the set of personalized models: the idea is that two neighboring agents should agree on each other's models.

\textbf{Problem reformulation.}
% Let us denote by $|\Nei{i}|=|\Nei{i}|$ the number of neighbors of agent $i$.
Let $\Theta_{i}$ be the set of $|\Nei{i}|+1$ variables  $\theta_j\in\mathbb{R}^p$ for $j \in \Nei i \cup \{i\}$, and denote $\theta_j$ by $\Theta_i^j$. This is similar to the notations used in Section~\ref{sec:modelpropagation}, except that here we consider $\Theta_i$ as living in $\R^{(|\Nei{i}|+1)\times p}$. We now define
$$\costcol^i(\Theta_{i})=\frac{1}{2}\sum_{j\in\Nei i}W_{ij} {\lVert \theta_i - \theta_j \rVert}^2 + \mu  D_{ii} \Loss_i(\theta_i),$$
so that we can rewrite our problem \eqref{eq:Qcl} as $\min_{\Theta\in\R^{n\times p}} \sum_{i=1}^n \costcol^i(\Theta_i)$.
% \begin{equation}\label{eq:mincl}
% \min_{\Theta\in\R^{n\times p}} \sum_{i=1}^n \costcol^i(\Theta_i).
% \end{equation}

% Let us rewrite
% \begin{align*}
%     \label{eq:Qcl}
%     \costcol(\Theta) &=\sum_{i=1}^n \bigg(\frac{1}{2}\sum_{j\in\Nei i}W_{ij} {\lVert \theta_i - \theta_j \rVert}^2 + \mu  D_{ii} \Loss_i(\theta_i)\bigg)\\
%   & :=  \sum_{i=1}^n \costcol^i(\Theta_{i})
% \end{align*}
In this formulation, the objective functions associated with the agents are
dependent as they share some decision variables in $\Theta$. In order to apply decentralized ADMM, we need to decouple the objectives. The idea is to introduce a local copy $\ttheta_i\in\R^{(|\Nei{i}|+1)\times p}$ of the decision
variables $\Theta_i$ for each agent $i$ and to impose
equality constraints on the variables $\ttheta^i_i=\ttheta^i_j$ for all $i\in\intset{n}, j\in\Nei{i}$. This partial consensus can be seen as requiring that two neighboring agents agree on each other's personalized model. We further introduce 4 secondary variables $Z^i_{ei}$, $Z^i_{ej}$, $Z^j_{ei}$ and $Z^j_{ej}$ for each edge $e=(i,j)$, which can be viewed as estimates of
the models $\ttheta_i$ and $\ttheta_j$ known by each end of $e$ and will allow an efficient decomposition of the ADMM updates.

Formally, denoting $\ttheta=[\ttheta_1^\top,\dots,\ttheta_n^\top]^\top\in \R^{(2|E|+n)\times p}$ and $Z\in\R^{4|E|\times p}$, we introduce the formulation
\begin{equation}
\label{eq:clz}
\begin{aligned}
&\min_{\substack{\ttheta\in\R^{(2|E|+n)\times p}\\ Z\in \mathcal{C}_E}} && \sum_{i=1}^n \costcol^i(\ttheta_i)\\
&\text{s.t. } \forall e=(i,j)\in E, && \left \{
  \begin{aligned}
  &Z^i_{ei}=\ttheta^i_i, \ Z^j_{ei}=\ttheta^j_i\\
  &Z^j_{ej}=\ttheta^j_j, \ Z^i_{ej}=\ttheta^i_j,
  \end{aligned}
  \right.
\end{aligned}
\end{equation}
where $\mathcal{C}_E=\{Z\in\R^{4|E|\times p}\mid  Z^i_{ei}=Z^i_{ej}, Z^j_{ej}=Z^j_{ei}$ for all $e=(i,j)\in E\}$. It is easy to see that Problem~\eqref{eq:clz} is equivalent to the original problem \eqref{eq:Qcl} in the following sense: the minimizer $\ttheta^{\star}$ of \eqref{eq:clz} satisfies $(\ttheta^\star)_i^j=\Theta_j^\star$ for all $i\in\intset{n}$ and $j\in\Nei{i}\cup\{i\}$.
Further observe that the set of constraints involving $\ttheta$ can be written $D\ttheta+HZ=0$ where $H=-I$ of dimension $4|E|\times 4|E|$ is diagonal invertible and $D$ of dimension $4|E|\times (2|E|+n)$ contains exactly one entry of 1 in each row. The assumptions of \cite{Wei2013a} are thus met and we can apply asynchronous decentralized ADMM.

Before presenting the algorithm, we derive the augmented Lagrangian associated with Problem~\eqref{eq:clz}. Let $\Lambda_{ei}^j$ be dual variables associated with constraints involving $\ttheta$ in~\eqref{eq:clz}. For convenience, we denote by $Z_i\in\R^{2|\Nei{i}|}$ the set of secondary variables ${\{\{Z_{ei}^i\} \cup \{Z_{ei}^j\}\}_{e=(i,j)\in E}}$ associated with agent $i$. Similarly, we denote by $\Lambda_i\in\R^{2|\Nei{i}|}$ the set of dual variables ${\{\{\Lambda_{ei}^i\} \cup \{\Lambda_{ei}^j\}\}_{e=(i,j)\in E}}$. The augmented Lagrangian is given by:
\begin{equation*}
    L_\rho(\ttheta, Z, \Lambda) = \sum_{i=1}^n L_\rho^i(\ttheta_i, Z_{i}, \Lambda_{i}),
\end{equation*}
where $\rho>0$ is a penalty parameter, $Z\in\mathcal{C}_E$ and
\begin{multline*}
    L_\rho^i(\ttheta_i, Z_{i}, \Lambda_{i}) = \costcol^i(\ttheta_i) + \sum_{j : e=(i,j)\in E} \Big[ \Lambda_{ei}^i (\ttheta_i^i - Z_{ei}^i)\\ + \Lambda_{ei}^j (\ttheta_i^j - Z_{ei}^j) + \frac{\rho}{2} \left( \lVert {\ttheta_i^i - Z_{ei}^i \rVert}^2 + \lVert {\ttheta_i^j - Z_{ei}^j \rVert}^2 \right)\Big].
\end{multline*}

\textbf{Algorithm.} ADMM consists in approximately minimizing the augmented Lagrangian $L_\rho(\ttheta, Z, \Lambda)$ by alternating minimization with respect to the primal variable $\ttheta$ and the secondary variable $Z$, together with an iterative update of the dual variable $\Lambda$.

We first briefly discuss how to instantiate the initial values $\ttheta(0)$, $Z(0)$ and $\Lambda(0)$. The only constraint on these initial values is to have $Z(0)\in\mathcal{C}_E$, so a simple option is to initialize all variables to $0$. That said, it is typically advantageous to use a warm-start strategy. For instance, each agent $i$ can send its solitary model $\theta_i^\loc$ to its neighbors, and then set $\ttheta_i^i=\theta_i^\loc$, $\ttheta_i^j=\theta_j^\loc$ for all $j\in\Nei{i}$, $Z_{ei}^i = \ttheta_i^i$, $Z_{ei}^j = \ttheta_i^j$ for all $e=(i,j)\in E$, and $\Lambda(0)=0$. Alternatively, one can initialize the algorithm with the model propagation solution obtained using the method of Section~\ref{sec:modelpropagation}.

Recall from Section~\ref{ssec:asyncmp} that in the asynchronous setting, a single agent wakes up at each time step and selects one of its neighbors. Assume that agent $i$ wakes up at some iteration $t\geq 0$ and selects $j\in\Nei{i}$. Denoting $e=(i,j)$, the iteration goes as follows:
\begin{enumerate}
\item Agent $i$ updates its primal variables:
  \begin{equation*}
      \ttheta_i(t+1) = \argmin_{\Theta \in \R^{(|\Nei{i}|+1) \times p}} L_\rho^i(\Theta, Z_{i}(t), \Lambda_{i}(t)), 
  \end{equation*}
  and sends $\ttheta_i^i(t+1), \ttheta_i^j(t+1), \Lambda_{ei}^i(t), \Lambda_{ei}^j(t)$ to agent $j$. Agent $j$ executes the same steps w.r.t. $i$.
\item  Using $\ttheta_j^j(t+1), \ttheta_j^i(t+1), \Lambda_{ej}^j(t), \Lambda_{ej}^i(t)$ received from $j$, agent $i$ updates its secondary variables:
\begin{gather*}
  \begin{split}
    Z_{ei}^i(t+1) = 
    \frac{1}{2}\Bigr [ \frac{1}{\rho}\big(\Lambda_{ei}^i(t)+\Lambda_{ej}^i(t)\big)& + \\[-1em]
    \ttheta_i^i(t+1)&+ \ttheta_j^i(t+1) \Bigl],
  \end{split}\\
  \begin{split}
    Z_{ei}^j(t+1) = 
    \frac{1}{2}\Bigr[ \frac{1}{\rho}\big(\Lambda_{ej}^j(t)+\Lambda_{ei}^j(t)\big)& + \\[-1em] 
    \ttheta_j^j(t+1)&+ \ttheta_i^j(t+1) \Bigl]. 
  \end{split}
\end{gather*}
  Agent $j$ updates its secondary variables symmetrically, so by construction we have $Z(t+1)\in\mathcal{C}_E$.
\item Agent $i$ updates its dual variables:
  \begin{equation*}
    \begin{aligned}
      \Lambda_{ei}^i(t+1) &= \Lambda_{ei}^i(t) + \rho\big(\ttheta_i^i(t+1) - Z_{ei}^i(t+1)\big), \\
      \Lambda_{ei}^j(t+1) &= \Lambda_{ei}^j(t) + \rho\big(\ttheta_i^j(t+1) - Z_{ei}^j(t+1)\big).
    \end{aligned}
  \end{equation*}
  Agent $j$ updates its dual variables symmetrically.
\end{enumerate}
All other variables in the network remain unchanged.

Step 1 has a simple solution for some loss functions commonly used in machine learning (such as quadratic and $L_1$ loss), and when it is not the case ADMM is typically robust to approximate solutions to the corresponding subproblems (obtained for instance after a few steps of gradient descent), see \citet{boyd2011distributed} for examples and further practical considerations.
Asynchronous ADMM converges almost surely to an optimal solution at a rate of $O(1/t)$ for convex objective functions \citep[see][]{Wei2013a}.

%%% Local Variables:
%%% mode: latex
%%% ispell-local-dictionary: "english"
%%% TeX-master: "main"
%%% End:

% !TEX root = main.tex

\section{Experiments}
\label{sec:experiments}

In this section, we provide numerical experiments to evaluate the performance of our decentralized algorithms with respect to accuracy, convergence rate and the amount of communication. %We study the synchronous and asynchronous approaches in both algorithms of collaborative learning and model propagation.
To this end, we  introduce two synthetic collaborative tasks: mean estimation and linear classification.

% In this section, we provide numerical experiments to show the relevance of the proposed model propagation and collaborative learning formulations. We also evaluate the performance of our decentralized algorithms with respect to the amount of communication.
% To this end, we will introduce two novel collaborative tasks: mean estimation and linear classification.
% \marc{Intro plus précise sur les objectifs: intérêt de l'introduction des valeurs de confiance ; vitesse de convergence en fonction du nombre de communications en synchrone ou asynchrone, ...}
% \marc{Uniformiser erreur ou accuracy}
% \marc{Mettre à dispo les codes des expés}

\subsection{Collaborative Mean Estimation}

\begin{figure*}[t]
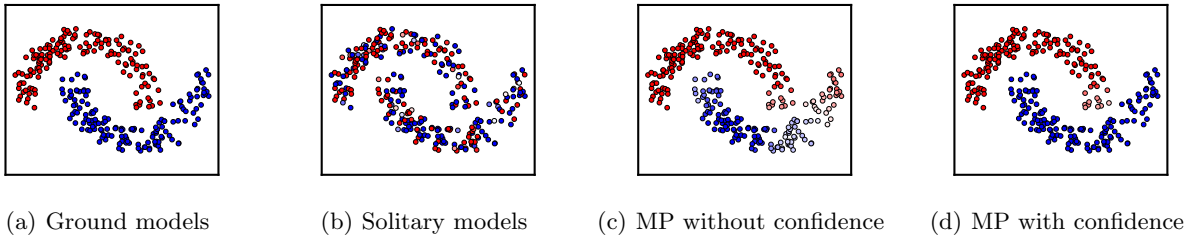

    \centering
    \vskip-1em
    \subfigure[Ground models \label{fig:mp_layout1}]{\hspace*{.33cm}\resizebox{0.2\textwidth}{!}{\input{images/mp/exp0/fig1.pgf}}\hspace*{.33cm}}
    \subfigure[Solitary models \label{fig:mp_layout2}]{\hspace*{.33cm}\resizebox{0.2\textwidth}{!}{\input{images/mp/exp0/fig2.pgf}}\hspace*{.33cm}}
    \subfigure[MP without confidence \label{fig:mp_layout3}]{\hspace*{.33cm}\resizebox{0.2\textwidth}{!}{\input{images/mp/exp0/fig3.pgf}}\hspace*{.33cm}}
    \subfigure[MP with confidence \label{fig:mp_layout4}]{\hspace*{.33cm}\resizebox{0.2\textwidth}{!}{\input{images/mp/exp0/fig4.pgf}}\hspace*{.33cm}}
    \caption{Illustration of the collaborative mean estimation task, where each point represents an agent and its 2D coordinates the associated auxiliary information. Figure~\ref{fig:mp_layout1} shows the ground truth models (blue for mean 1 and red for mean -1). Figure~\ref{fig:mp_layout2} shows the solitary models (local averages) for an instance where $\epsilon=1$. Figures~\ref{fig:mp_layout3}-\ref{fig:mp_layout4} show the models after propagation, without/with the use of confidence values.}
    \label{fig:mp_exp_layout}
\end{figure*}

We first introduce a simple task in which the goal of each agent is to estimate the mean of a 1D distribution. To this end, we adapt the two intertwining moons dataset popular in semi-supervised learning \citep{zhou2004learning}. We consider a set of $300$ agents, together with auxiliary information about each agent $i$ in the form of a vector $v_i\in\mathbb{R}^2$. The true distribution $\mu_i$ of an agent $i$ is either $\mathcal{N}(1,40)$ or $\mathcal{N}(-1,40)$ depending on whether $v_i$ belongs to the upper or lower moon, see Figure~\ref{fig:mp_layout1}.
Each agent $i$ receives $m_i$ samples $x_i^1,\dots,x_i^{m_i}\in\mathbb{R}$ from its distribution $\mu_i$. Its solitary model is then given by $\theta_i^{\loc} = \frac{1}{m_i} \sum_{j=1}^{m_i} x_i^j$, which corresponds to the use of the quadratic loss function $\loss(\theta; x_i) = {\lVert \theta - x_i \rVert}^2$. Finally, the graph over agents is the complete graph where the weight between agents $i$ and $j$ is given by a Gaussian kernel on the agents' auxiliary information
${W_{ij} = \exp ( {-{\lVert v_i - v_j \rVert}^2}/{2\sigma^2} )},$
with $\sigma=0.1$ for appropriate scaling. In all experiments, the parameter $\alpha$ of model propagation was set to $0.99$, which gave the best results on a held-out set of random problem instances.
We first use this mean estimation task to illustrate the importance of considering confidence values in our model propagation formulation, and then to evaluate the efficiency of our asynchronous decentralized algorithm.

\textbf{Relevance of confidence values.}
Our goal here is to show that introducing confidence values into the model propagation approach can significantly improve the overall accuracy, especially when the agents receive unbalanced amounts of data. In this experiment, we only compare model propagation with and without confidence values, so we compute the optimal solutions directly using the closed-form solution \eqref{eq:mp_closedform}.

\begin{figure*}[t]
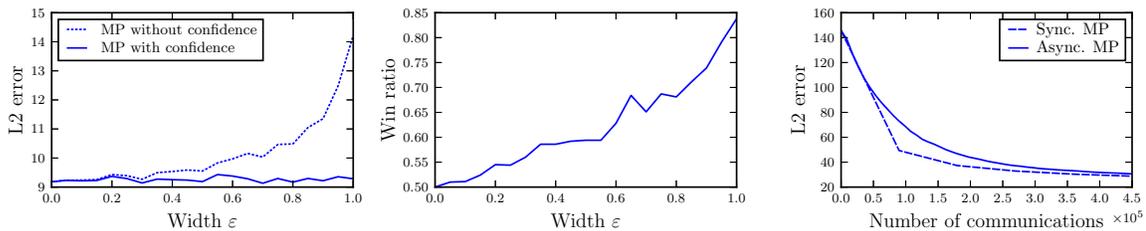

  \centering
      \vskip-.5em
  \subfigure%[\label{fig:mp_exp1}]
  {\resizebox{0.6\textwidth}{!}{\input{images/mp/exp1/fig.pgf}}}
  \subfigure%[\label{fig:mp_exp2}]
  {\resizebox{0.3\textwidth}{!}{\input{images/mp/exp2/fig.pgf}}}
      \vskip-.5em
  \caption{Results on the mean estimation task.
  %Comparison of the model propagation approach with and without the use of confidence values depending on the unbalancedness of the local dataset sizes (collaborative mean estimation task).
  (Left-middle) Model propagation with and without confidence values w.r.t. the unbalancedness of the local datasets. The left figure shows the $L_2$ errors, while the middle one shows the percentage of wins in favor of using confidence values. (Right) $L_2$ error of the synchronous and asynchronous model propagation algorithms with respect to the number of pairwise communications.}
    \label{fig:mp}
\end{figure*}

We generate several problem instances with varying standard deviation for the confidence values $c_i$'s. More precisely, we sample $c_i$ for each agent $i$ from a uniform distribution centered at $1/2$ with width $\epsilon\in[0, 1]$. The number of samples $m_i$ given to agent $i$ is then set to ${m_i = \lceil c_i \cdot 100 \rceil}$. The larger $\epsilon$, the more variance in the size of the local datasets. Figures~\ref{fig:mp_layout2}-\ref{fig:mp_layout4} give a visualization of the models before and after propagation on a problem instance for the hardest setting $\epsilon=1$.
Figure~\ref{fig:mp} (left-middle) shows results averaged over $1000$ random problem instances for several values of $\epsilon$. As expected, when the local dataset sizes are well-balanced (small $\epsilon$), model propagation performs the same with or without the use of confidence values. Indeed, both have similar $L_2$ error with respect to the target mean, and the win ratio is about $0.5$. However, the performance gap in favor of using confidence values increases sharply with $\epsilon$. For $\epsilon=1$, the win ratio in favor of using confidence values is about $0.85$. Strikingly, the error of model propagation with confidence values remains constant as $\epsilon$ increases. These results empirically confirm the relevance of introducing confidence values into the objective function.

\begin{figure*}[t]
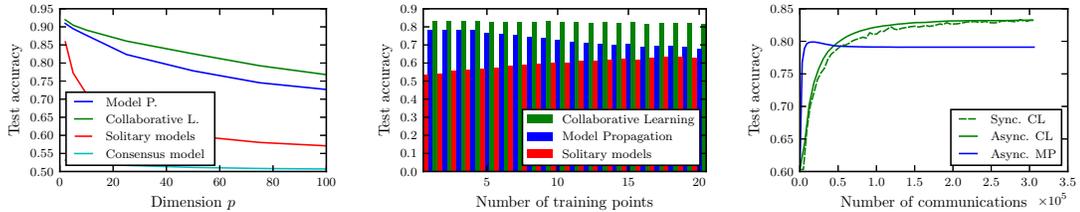

    \centering
    \vskip-1em
    \subfigure%[\label{fig:cl_exp1}]
    {\resizebox{0.28\textwidth}{!}{\input{images/cl/exp1/fig.pgf}}}
    \subfigure%[\label{fig:cl_exp3}]
    {\resizebox{0.28\textwidth}{!}{\input{images/cl/exp3/fig.pgf}}}
    \subfigure%[\label{fig:cl_exp2}]
    {\resizebox{0.28\textwidth}{!}{\input{images/cl/exp2/fig.pgf}}}
      \vskip-.5em
    \caption{Results on the linear classification task. (Left) Test accuracy of model propagation and collaborative learning with varying feature space dimension. (Middle) Average test accuracy of model propagation and collaborative learning with respect to the number of training points available to the agent (feature dimension $p=50$). (Right) Test accuracy of synchronous and asynchronous collaborative learning and asynchronous model propagation with respect to the number of pairwise communications (linear classification task, $p=50$).}
    \label{fig:cl}
\end{figure*}

\textbf{Asynchronous algorithm.}
% \begin{figure}[t]
%     \centering
%     \input{images/mp/exp2/fig.pgf}
%     \caption{$L_2$ error of the synchronous and asynchronous model propagation algorithms with respect to the number of pairwise communications (collaborative mean estimation task).}
%     \label{fig:mp_exp2}
% \end{figure}
In this second experiment, we compare asynchronous model propagation with the synchronous variant given by \eqref{eq:modelit}. We are interested in the average $L_2$ error of the models as a function of the number of pairwise communications (number of exchanges from one agent to another). Note that a single iteration of the synchronous (resp. asynchronous) algorithm corresponds to $2|E|$ (resp. $2$) communications.
For the asynchronous algorithm, we set the neighbor selection distribution $\pi_i$ of agent $i\in\intset{n}$ to be uniform over the set of neighbors $\mathcal{N}_i$.

Figure~\ref{fig:mp} (right) shows the results on a problem instance generated as in the previous experiment (with $\varepsilon = 1$). Since the asynchronous algorithm is randomized, we average its results on $100$ random runs. We see that our asynchronous algorithm achieves an accuracy/communication trade-off which is almost as good as that of the synchronous one, without requiring any synchronization. It is thus expected to be much faster than the synchronous algorithm on large decentralized networks with communication delays and/or without efficient global synchronization.

\subsection{Collaborative Linear Classification}
\label{ssec:exp_cl}

In the previous mean estimation task, the squared distance between two model parameters (i.e., estimated means) translates into the same difference in $L_2$ error with respect to the target mean. Therefore, our collaborative learning formulation is essentially equivalent to our model propagation approach. To show the benefits that can be brought by collaborative learning, we now consider a linear classification task. Since two linear separators with significantly different parameters can lead to similar predictions on a given dataset, incorporating the local errors into the objective function rather than simply the distances between parameters should lead to more accurate models.

% \begin{figure}[t]
%     \centering
%     {\resizebox{0.32\textwidth}{!}{\input{images/cl/exp0/fig.pgf}}}
%     \caption{Target models of the agents (represented as points in $\mathbb{R}^2$) for the collaborative linear classification task. Two models are linked together when the angle between them is small, which corresponds to a small distance after projection onto the unit circle.}
%     \label{fig:cl_exp0}
% \end{figure}

We consider a set of $100$ agents whose goal is to perform linear classification in $\mathbb{R}^p$. For ease of visualization, the target (true) model of each agent lies in a 2-dimensional subspace: we represent it as a vector in $\mathbb{R}^p$ with the first two entries drawn from a normal distribution centered at the origin and the remaining ones equal to $0$. We consider  the similarity graph where the weight between two agents $i$ and $j$ is a Gaussian kernel on the distance between target models, where the distance here refers to the length of the chord of the angle $\phi_{ij}$ between target models projected on a unit circle. More formally, $W_{i,j}=\exp((\cos(\phi_{i,j})-1)/\sigma)$ with $\sigma = 0.1$ for appropriate scaling. Edges with negligible weights are ignored to speed up computation. We refer the reader to \suppapp{the supplementary material}{\ref{app:exp}} for a 2D visualization of the target models and the links between them.
Every agent receives a random number of training points drawn uniformly between $1$ and $20$. Each training point (in $\mathbb{R}^p$) is drawn uniformly around the origin, and the binary label is given by the prediction of the target linear separator. We then add some label noise by randomly flipping each label with probability $0.05$.
The loss function used by the agents is the hinge loss, given by $\loss(\theta; (x_i,y_i)) = \max \big( 0, 1 - y_i \theta\T x_i \big)$.
% used in SVM:
% \begin{equation*}
%     \loss(\theta; (x_i,y_i)) = \max \left( 0, 1 - y_i \theta\T x_i \right).
% \end{equation*}
As in the previous experiment, for each algorithm we tune the value of $\alpha$ on a held-out set of random problem instances. Finally, we will evaluate the quality of the learned model of each agent by computing the accuracy on a separate sample of $100$ test points drawn from the same distribution as the training set.

In the following, we use this linear classification task to compare the performance of collaborative learning against model propagation, and to evaluate the efficiency of our asynchronous algorithms. 

\textbf{MP vs. CL.}
% \begin{figure}[t]
%     \centering
%     \input{images/cl/exp1/fig.pgf}
%     \caption{Test accuracy of model propagation and collaborative learning on the linear classification task with varying feature space dimension.}
%     \label{fig:cl_exp1}
% \end{figure}
In this first experiment, we compare the accuracy of the models learned by model propagation and collaborative learning with feature space dimension $p$ ranging from $2$ to $100$. Figure~\ref{fig:cl} (left) shows the results averaged over $10$ randomly generated problem instances for each value of $p$. As baselines, we also plot the average accuracy of the solitary models and of the global consensus model minimizing \eqref{eq:consensus}. The accuracy of all models decreases with the feature space dimension, which comes from the fact that the expected number of training samples remains constant. As expected, the consensus model achieves very poor performance since agents have very different objectives. On the other hand, both model propagation and collaborative learning are able to improve very significantly over the solitary models, even in higher dimensions where on average these initial models barely outperform a random guess. Furthermore, collaborative learning always outperforms model propagation. %Note that the performance gap tends to increase when the dimension grows. This is because the models have more degrees of freedom, hence committing to a small region around the parameters of the initial models becomes very restrictive.

We further analyze these results by plotting the accuracy with respect to the size of the local training set (Figure~\ref{fig:cl}, middle). As expected, the accuracy of the solitary models is higher for larger training sets. Furthermore, collaborative learning converges to models which have similar accuracy regardless of the training size, effectively correcting for the initial unbalancedness. While model propagation also performs well, it is consistently outperformed by collaborative learning on all training sizes. This gap is larger for agents with more training data: in model propagation, the large confidence values associated with these agents prevent them from deviating much from their solitary model, thereby limiting their own gain in accuracy. %, for which model propagation provides less improvement.

\textbf{Asynchronous algorithms.}
% \begin{figure}[t]
%     \centering
%     \input{images/cl/exp2/fig.pgf}
%     \caption{Test accuracy of synchronous and asynchronous collaborative learning and asynchronous model propagation with respect to the number of pairwise communications (linear classification task, $p=50$).}
%     \label{fig:cl_exp2}
% \end{figure}
This second experiment compares our asynchronous collaborative learning algorithm with a synchronous variant also based on ADMM (see \suppapp{supplementary material}{\ref{app:syncADMM}} for details) in terms of number pairwise of communications.
Figure~\ref{fig:cl} (right) shows that our asynchronous algorithm performs as good as its synchronous counterpart and should thus be largely preferred for deployment in real peer-to-peer networks. It is also worth noting that asynchronous model propagation converges an order of magnitude faster than collaborative learning, as it only propagates models that are pre-trained locally.
Model propagation can thus provide a valuable warm-start initialization for collaborative learning.

\textbf{Scalability.} We also observe experimentally that the number of iterations needed by our decentralized algorithms to converge scales favorably with the size of the network (see \suppapp{supplementary material}{\ref{app:exp}} for details).

%%% Local Variables:
%%% mode: latex
%%% ispell-local-dictionary: "english"
%%% TeX-master: "main"
%%% End:

% !TEX root = main.tex

\section{Conclusion}
\label{sec:conclu}

We proposed, analyzed and evaluated two asynchronous peer-to-peer algorithms for the novel setting of decentralized collaborative learning of personalized models.
% In this paper, we proposed novel contributions for the decentralized peer-to-peer machine learning paradigm. While most of the work on big data has focused on very large centralized datasets (or distributed on master/slaves commodity clusters), the originality of our setting is in contrast to consider many learners with personalized objectives and small datasets organized in a decentralized network. In this context, we proposed two approaches: a model propagation method which incorporates confidence values, and a collaborative learning method which trades off between local error and network smoothness of the learned models. For both approaches, we have proposed and analyzed synchronous and asynchronous algorithms which allow the learning phase to be performed in a fully decentralized manner.
% Finally, we studied the empirical performance of the proposed methods on two novel collaborative learning tasks.
This work opens up interesting perspectives. %First, while we proved that our asynchronous algorithm for model propagation converges to the desired solution, we aim to study its convergence rate. We expect it to be linear, similar to existing gossip algorithms for consensus averaging \citep{boyd2006randomized}.
The link between the similarity graph and the generalization performance of the resulting models should be formally analyzed. This could in turn guide the design of generic methods to estimate the graph weights, making our approaches widely applicable. Other directions of interest include the development of privacy-preserving algorithms as well as extensions to time-evolving networks and sequential arrival of data.

% We also hope to develop online versions of our decentralized algorithms for collaborative learning both in the active and passive learning settings, and to analyze their performance by establishing theoretical guarantees in the form of mistake bounds.

%%% Local Variables:
%%% mode: latex
%%% ispell-local-dictionary: "english"
%%% TeX-master: "main"
%%% End:

\paragraph{Acknowledgments} This work was partially supported by grant ANR-16-CE23-0016-01 and by a grant from CPER Nord-Pas de Calais/FEDER DATA Advanced data science and technologies 2015-2020.

\clearpage
\bibliographystyle{apalike}
\bibliography{main}

\onecolumn
\newpage
\appendix

\renewcommand{\thesection}{Appendix~\Alph{section}}
% !TEX root = supplementary.tex

\section{Proof of Proposition~\ref{prop:MPclosedform}}
\label{app:prop1}
\begin{repproposition}{prop:MPclosedform}[Closed-form solution]
  % \label{prop:MPclosedform}
  Let $P = D^{-1}W$ be the stochastic similarity matrix associated
  with the graph $G$ and
  $\Theta^{\loc} =
  [\theta^{\loc}_1;\dots;\theta^{\loc}_n]\in\mathbb{R}^{n\times
    p}$.
  The solution
  ${\Theta^\star = \argmin_{\Theta \in \mathbb{R}^{n \times p}}
    \costmp(\Theta)}$ is given by
  \begin{equation*}
    \Theta^\star = \balpha {(I - \balpha (I-C) - \alpha P)}^{-1} C\Theta^{\loc},
  \end{equation*}
  with $\alpha\in(0,1)$ such that $\mu = \balpha  / \alpha$, and $\balpha=1-\alpha$.
\end{repproposition}

\begin{proof}
  We write the objective function in matrix form:
  \begin{equation*}
    \costmp(\Theta) = \frac{1}{2} \left(\tr{[{\Theta}\T L {\Theta}]} + \mu \tr{[{(\Theta - \Theta^{\loc})}\T DC {(\Theta - \Theta^{\loc})}]} \right),
  \end{equation*}
  where $L= D - W$ is the graph Laplacian matrix and $\tr$ denotes the
  trace of a matrix. As $\costmp(\Theta)$ is convex and quadratic in
  $\Theta$, we can find its global minimum by setting its derivative
  to
  0. %Since $\nabla \costmp(\Theta) = L \Theta + \mu DC (\Theta - \Theta^{\loc})$, we have:
  \begin{align*}
    \nabla \costmp(\Theta) &= L \Theta + \mu DC (\Theta - \Theta^{\loc})\\
                           & = L \Theta^* + \mu DC (\Theta^* - \Theta^{\loc}) \\
                           &= (D - W + \mu DC) \Theta^* - \mu DC \Theta^{\loc}.
  \end{align*}
  Hence,
  \begin{align*}
    \nabla \costmp(\Theta) =  0 & \Leftrightarrow (I - P + \mu C) \Theta^* - \mu C \Theta^{\loc}=0\\
                                &\Leftrightarrow (I - \balpha (I-C) - \alpha P)\Theta^* - \balpha C\Theta^{\loc}=0,
  \end{align*}
  with $\mu = \balpha  / \alpha$. Since $P$ is a stochastic
  matrix, its eigenvalues are in $[-1, 1]$. Moreover,
  ${ {(I-C)}_{ii} < 1 }$ for all $i$, thus
  ${\rho(\balpha  (I-C) + \alpha P) < 1}$ where $\rho(\cdot)$
  denotes the spectral radius. Consequently,
  $I - \balpha (I-C) - \alpha P$ is invertible and we get the
  desired result.
\end{proof}
\section{Convergence of the Iterative Form \eqref{eq:modelit}}
\label{app:iter}

We can rewrite the equation
\begin{equation}
    \Theta(t+1) = {(\alpha I + \balpha  C)}^{-1} \left(\alpha P \Theta(t) + \balpha C\Theta^{\loc}\right),
    \tag{\ref{eq:modelit}}
\end{equation}
as 
\begin{equation*}
        \Theta(t) = {\left({(\alpha I + \balpha  C)}^{-1} \alpha P\right)}^t \Theta(0) + \sum_{k=0}^{t-1} {\left({(\alpha I + \balpha  C)}^{-1} \alpha P\right)}^k {(\alpha I + \balpha  C)}^{-1} \balpha C\Theta^{\loc}.
\end{equation*}
%Recall from Appendix~\ref{app:prop1} that ${\rho(\balpha  (I-C) + \alpha P) < 1}$.
Since $\frac{\alpha}{(\alpha + \balpha c_i)} < 1$ for any $i \in \intset{n}$, we have ${\rho\left({(\alpha I + \balpha  C)}^{-1} \alpha P\right) < 1}$ and therefore:
 $$\lim\limits_{t \rightarrow \infty} {\left({(\alpha I + \balpha  C)}^{-1} \alpha P\right)}^t = 0,$$
hence
\begin{align*}
    \lim\limits_{t \rightarrow \infty} \Theta(t)
    = {}& {\left(I - {(\alpha I + \balpha  C)}^{-1} \alpha P\right)}^{-1} {(\alpha I + \balpha  C)}^{-1} \balpha C\Theta^{\loc} \\
    %= {}& {\left( \alpha I +\balpha C - \alpha P \right)}^{-1} \balpha  C \Theta^{\loc} \\
    = {}& {\left( I - \balpha (I-C) - \alpha P \right)}^{-1} \balpha  C \Theta^{\loc} \\
    = {}& \Theta^*.
\end{align*}

\section{Proof of Theorem~\ref{thm:asyncMP}}
\label{app:async_mp}
%\subsection{Convergence Analysis}

\begin{reptheorem}{thm:asyncMP}[Convergence]
Let $\ttheta(0)\in\mathbb{R}^{n^2\times p}$ be some arbitrary initial value and $(\ttheta(t))_{t\in\mathbb{N}}$ be the sequence generated by our model propagation algorithm. Let $ \Theta^\star = \argmin_{\Theta \in \mathbb{R}^{n \times p}} \costmp(\Theta)$ be the optimal solution to model propagation. For any $i\in\intset{n}$, we have:
$$\lim\limits_{t \rightarrow \infty} \mathbb{E} \left[ \ttheta_i^j(t) \right] =  \Theta^\star_j \text{ for } j \in \mathcal{N}_i \cup \{i\}.$$
\end{reptheorem}

\begin{proof}
In order to prove the convergence of our algorithm, we need to introduce an equivalent formulation as a random iterative process over $\ttheta\in\mathbb{R}^{n^2\times p}$, the horizontal stacking of all the $\ttheta_i$'s.

The \emph{communication step} of agent $i$ with its neighbor $j$ consists in overwriting $\tilde \Theta_i^j$ and $\tilde \Theta_j^i$ with respectively $\tilde \Theta_j^j$ and $\tilde \Theta_i^i$. This step will be handled by multiplication with the matrix $O(i,j)\in\mathbb{R}^{n^2\times n^2}$ defined as
$$O(i,j) = I + e_i^j(e_j^j-e_i^j)\T + e_j^i(e_i^i-e_j^i)\T,$$
where for $i,j\in\intset{n}$, the vector $e_i^j\in\mathbb{R}^{n^2}$ has 1 as its $(i-1)n+j$-th coordinate and 0 in all others.

%The \emph{update step} of node $i$ consists in replacing $\tilde \Theta_i^i$ with the $i$-th line of ${(\alpha I + \balpha  C)}^{-1} \big( \alpha P \ttheta_i + \balpha C\Theta^{\loc} \big)$. This step will be handled by multiplication with the matrix $U(i)\in\mathbb{R}^{n^2\times n^2}$ and addition of the vector $u(i)\in\mathbb{R}^{n^2 \times p}$ defined as follows:
The \emph{update step} of node $i$ and $j$ consists in replacing $\ttheta_i^i$ and $\ttheta_j^j$ with respectively the $i$-th line of ${(\alpha I + \balpha  C)}^{-1} \big( \alpha P \ttheta_i + \balpha C\Theta^{\loc} \big)$ and the $j$-th line of ${(\alpha I + \balpha  C)}^{-1} \big( \alpha P \ttheta_j + \balpha C\Theta^{\loc} \big)$. This step will be handled by multiplication with the matrix $U(i,j)\in\mathbb{R}^{n^2\times n^2}$ and addition of the vector $u(i,j)\in\mathbb{R}^{n^2 \times p}$ defined as follows:
\begin{align*}
    U(i,j) &= I + (e_i^ie_i^{i\top} + e_j^je_j^{j\top})(M-I)\\
    u(i,j) &= (e_i^ie_i^{i\top} + e_j^je_j^{j\top}) {(\alpha I + \balpha  C)}^{-1} \balpha  C \ttheta^{\loc} ,
\end{align*}
where $M\in\mathbb{R}^{n^2\times n^2}$ is a block diagonal matrix with repetitions of ${(\alpha I + \balpha  C)}^{-1} \alpha P$ on the diagonal and $\ttheta^{\loc} \in \R^{n^2 \times p}$ is built by stacking horizontally $n$ times the matrix $\Theta^{\loc}$.

We can now write down a global iterative process which is equivalent to our model propagation algorithm. For any $t\geq 0$:
\begin{equation*}
%\label{eq:globalalg}
\tilde\Theta(t+1) = A(t)\tilde\Theta(t) + b(t)
\end{equation*}
where,  
\begin{equation*}
  \left \{
  \begin{aligned}
    A(t) &= I^E U(i,j) O(i,j) \\
    b(t) &= u(i,j)
  \end{aligned}
  \right .\quad  \text{ w.p. } \frac{\pi_i^j}{n}
  \text{ for } i,j \in \llbracket 1,n \rrbracket,% \\
\end{equation*}
and $I^E$ is a $n^2 \times n^2$ diagonal matrix with its $(i-1)n+j$-th value equal to 1 if $(i,j) \in E$ or $i=j$ and equal to 0 otherwise.
Note that $I^E$ is used simply to simplify our analysis by setting to 0 the lines of $A(t)$ corresponding to non-existing edges (which can be safely ignored). 

% Also, it is possible to draw a parallel between the asynchronous algorithm and well known coordinate descent algorithms~\cite{wright2015coordinate}. While we could simply define an asynchronous algorithm inspired from randomized coordinate descent where the activated agent would contact all of its neighbors before updating its own value, we can see our gossip-typed algorithm as a randomized asynchronous coordinate descent with inconsistent reads and unbounded delay. The reason for not choosing to apply straightforwardly the results from~\cite{liu2015asynchronous} lies in the fact that experiments showed that our asynchronous algorithm performed better in general, which could be due to the fact that our cost function \eqref{eq:Qw} is quadratic in its parameters.

First, let us write the expected value of $\tilde\Theta(t)$ given $\tilde\Theta(t-1)$:
\begin{equation}
\label{eq:transition}
\mathbb{E} \left[ \tilde{\Theta}(t) | \tilde{\Theta}(t-1) \right] = \mathbb{E}[A(t)] \tilde{\Theta}(t-1) + \mathbb{E}[b(t)].
\end{equation}
Since the $A(t)$'s and $b(t)$'s are i.i.d., for any $t\geq 0$ we have $\bar A = \mathbb{E}[A(t)]$ and $\bar b = \mathbb{E}[b(t)]$ where
\begin{eqnarray*}
    \bar A &=& \frac{1}{n}I^E %\left(
    \sum_{i,j}^n \pi_i^j U(i,j)O(i,j), \\
    %+ \sum_{i=1}^n \pi_i^i U(i) \right),\\
    \bar b &=&
    \frac{1}{n}\sum_{i,j}^n \pi_i^j u(i,j).
\end{eqnarray*}

In order to prove Theorem~\ref{thm:asyncMP}, we first need to show that $\rho(\bar A) < 1$, where $\rho(\bar A)$ denotes the spectral radius of $\bar A$.
First, recall that ${\rho\left({(\alpha I + \balpha  C)}^{-1} \alpha P\right) < 1}$ (see Appendix~\ref{app:iter}). We thus have $\rho( M) < 1$ by construction of $M$ and
$$\lambda (I- M) \subset (0,2),$$
where $\lambda (\cdot)$ denotes the spectrum of a matrix.
Furthermore, from the properties in \cite{spectrumproduct} we know that
$$\lambda \left((e_i^ie_i^{i\top} + e_j^je_j^{j\top}) (I - M)\right) \subset [0,2],$$
and finally we have:
$$\lambda \left(I + (e_i^ie_i^{i\top} + e_j^je_j^{j\top}) ( M - I)\right) = \lambda (U(i,j)) \subset [-1,1].$$
As we also have $\lambda (O(i,j)) \subset [0,1]$ %and the properties in \eqref{eq:probasum},
therefore
$$\lambda \left(U(i,j)O(i,j)\right) \subset [-1,1].$$

Let us first suppose that $-1$ is an eigenvalue of $U(i,j)O(i,j)$ associated with the eigenvector $\tilde v$. From the previous inequalities we deduce that $\tilde v$ must be an eigenvector of $O(i,j)$ associated with the eigenvalue $+1$ and an eigenvector of $U(i,j)$ associated with the eigenvalue $-1$.
Then from $\tilde v = O(i,j) \tilde v$ we have $\tilde v_i^j = \tilde v_j^j$ and $\tilde v_j^i = \tilde v_i^i$. From $-\tilde v = U(i,j) \tilde v$ we can deduce that $\tilde v_k^l = 0$ for any $k \neq l$ or $k=l \in \intset{n} \backslash \{i,j\}$. Finally we can see that $\tilde v_i^i = \tilde v_j^j = 0$ and therefore $\tilde v = 0$. This proves by contradiction that $-1$ is not an eigenvalue of $U(i,j)O(i,j)$ and furthermore that $-1$ is not an eigenvalue of $\bar A$.

Let us now suppose that $+1$ is an eigenvalue of $\bar A$, associated with the eigenvector $\tilde v\in\mathbb{R}^{n^2}$. This would imply that
\begin{align*}
    \tilde v = \bar A \tilde v %\\
    = \frac{1}{n}I^E \sum_{i,j}^n \pi_i^j U(i,j)O(i,j) \tilde v.
%\Bigg( \sum_{i\neq j}^n \pi_i^j \Big( I + e_i^j{(e_j^j - e_i^j)}\T + e_j^i{(e_i^i - e_j^i)}\T \Big) \tilde v \\
%& \hspace{1.3cm} + \sum_{i=1}^n \pi_i^i (I + e_i^ie_i^{i\top} ( M - I))\tilde v \Bigg).
\end{align*}
This can be expressed line by line as the following set of equations:
\begin{align*}
    \sum_{k=1}^n (\pi_1^k + \pi_k^1) \tilde v_1^1 &= e_1^{1\top} \sum_{k=1}^n (\pi_1^k + \pi_k^1) MO(1,k) \tilde v \\
    (\pi_1^2 + \pi_2^1) \tilde v_1^2 &= (\pi_1^2 + \pi_2^1) \tilde v_2^2  & \text{ if } & (1,2)   \in E & \text{ else } &   \tilde v_1^2 = 0 \\
    (\pi_1^3 + \pi_3^1) \tilde v_1^3 &= (\pi_1^3 + \pi_3^1) \tilde v_3^3  & \text{ if } & (1,3)   \in E & \text{ else } &   \tilde v_1^3 = 0 \\
    & \vdotswithin{=}  \\
    (\pi_2^1 + \pi_1^2) \tilde v_2^1 &= (\pi_2^1 + \pi_1^2) \tilde v_1^1  & \text{ if } & (2,1)   \in E & \text{ else } &   \tilde v_2^1 = 0 \\
    \sum_{k=1}^n (\pi_2^k + \pi_k^2) \tilde v_2^2 &= e_2^{2\top} \sum_{k=1}^n (\pi_2^k + \pi_k^2) MO(2,k) \tilde v \\
    (\pi_2^3 + \pi_3^2) \tilde v_2^3 &= (\pi_2^3 + \pi_3^2) \tilde v_3^3  & \text{ if } & (2,3)   \in E & \text{ else } &   \tilde v_2^3 = 0 \\
    & \vdotswithin{=}  \\
    (\pi_n^{n-2} + \pi_{n-2}^n) \tilde v_n^{n-2} &= (\pi_n^{n-2} + \pi_{n-2}^n) \tilde v_{n-2}^{n-2}  & \text{ if } & (n,{n-2})   \in E & \text{ else } &   \tilde v_n^{n-2} = 0 \\
    (\pi_n^{n-1} + \pi_{n-1}^n) \tilde v_n^{n-1} &= (\pi_n^{n-1} + \pi_{n-1}^n) \tilde v_{n-1}^{n-1}  & \text{ if } & (n,{n-1})   \in E & \text{ else } &   \tilde v_n^{n-1} = 0 \\
    \sum_{k=1}^n (\pi_n^k + \pi_k^n) \tilde v_n^n &= e_n^{n\top} \sum_{k=1}^n (\pi_n^k + \pi_k^n) MO(n,k) \tilde v \\
\end{align*}
%\begin{align*}
    %{e_1^1}\T ( M - I) \tilde v &= 0 \\
    %{(e_2^2 - e_1^2)}\T \tilde v &= 0  & \text{ if } & (1,2)   \in E & \text{ else } &   \tilde v_1^2 = 0 \\
    %{(e_3^3 - e_1^3)}\T \tilde v &= 0  & \text{ if } & (1,3)   \in E & \text{ else } &   \tilde v_1^3 = 0 \\
    %& \vdotswithin{=}  \\
    %{(e_1^1 - e_2^1)}\T \tilde v &= 0  & \text{ if } & (2,1)   \in E & \text{ else } &   \tilde v_2^1 = 0 \\
    %{e_2^2}\T ( M - I) \tilde v &= 0 \\
    %{(e_3^3 - e_2^3)}\T \tilde v &= 0  & \text{ if } & (2,3)   \in E & \text{ else } &   \tilde v_2^3 = 0 \\
    %& \vdotswithin{=}  \\
    %{(e_{n-2}^{n-2} - e_{n}^{n-2})}\T \tilde v &= 0  & \text{ if } & (n,n-2) \in E & \text{ else } & \tilde v_n^{n-2} = 0 \\
    %{(e_{n-1}^{n-1} - e_{n}^{n-1})}\T \tilde v &= 0  & \text{ if } & (n,n-1) \in E & \text{ else } & \tilde v_{n}^{n-1} = 0 \\
    %{e_n^n}\T ( M - I) \tilde v &= 0.
%\end{align*}
We can rewrite the above system as
\begin{align}
\label{eq:system1}
\begin{split}
    \tilde v_i^j &=
    \begin{cases}
        v_j & \text{ if } (i,j) \in E \text{ or } i=j \\
        0 & \text{ otherwise,}
    \end{cases} \\
    0 &= \left(I - {(\alpha I + \balpha C)}^{-1} \alpha P\right) v.
\end{split}
\end{align}
with $v \in \mathbb{R}^{n \times p}$. As seen in Appendix~\ref{app:iter}, the matrix ${I - \balpha (I-C) - \alpha P}$ is invertible. Consequently $v = 0$ and thus $\tilde v = 0$, which proves by contradiction that $+1$ is not an eigenvalue of $\bar A$.

Now that we have shown that $\rho(\bar A) < 1$, let us write the expected value of $\tilde{\Theta}(t)$ by ``unrolling'' the recursion~\eqref{eq:transition}:
\begin{equation*}
\mathbb{E} \left[ \tilde{\Theta}(t) \right] = {\bar A}^t \tilde{\Theta}(0)
    + \sum_{k=0}^{t-1} {\bar A}^k \bar b.
\end{equation*}
Let us denote $\tilde{\Theta}^* = \lim\limits_{t \rightarrow \infty} \mathbb{E} \left[ \tilde \Theta(t) \right]$. Because $\rho(\bar A) < 1$, we can write
$$\tilde{\Theta}^*
    = {(I - \bar A)}^{-1} \bar b,$$
and finally
$$(I - \bar A) \tilde{\Theta}^* = \bar b.$$
Similarly as in \eqref{eq:system1}, we can identify $\hat\Theta \in \mathbb{R}^{n \times p}$ such that
\begin{align*}
    \tilde \Theta^{j*}_i &=
    \begin{cases}
         \hat\Theta_j & \text{ if } (i,j) \in E \text{ or } i=j \\
        0 & \text{ otherwise,}
    \end{cases} \\
    \balpha C\Theta^{\loc} &= (I - \balpha (I-C) + \alpha P) \hat\Theta.
\end{align*}
Recalling the results from Appendix~\ref{app:prop1}, we have
$$\hat\Theta = \balpha  {(I - \balpha (I-C) - \alpha P)}^{-1}C\Theta^{\loc},$$
and we thus have
$$\hat\Theta=\Theta^* = \argmin_{\Theta \in \mathbb{R}^{n \times p}} \costmp(\Theta),$$
and the theorem follows.
\end{proof}

\section{Synchronous Decentralized ADMM Algorithm for Collaborative Learning}
\label{app:syncADMM}

For completeness, we present here the \emph{synchronous} decentralized ADMM algorithm for
collaborative learning. Based on our reformulation of Section~\ref{ssec:async_cl} and following \citet{Wei2012a}, the algorithm to find $\ttheta^\star$ consists in iterating over the
following steps, starting at $t=0$:

\begin{enumerate}
    \item Every agent $i\in\intset{n}$ updates its primal variables:
        \begin{equation*} \ttheta_i(t+1) = \argmin_{\Theta \in \R^{(|\Nei{i}|+1) \times p}} L_\rho^i(\Theta, Z_{i}(t), \Lambda_{i}(t)),
    \end{equation*}
    and sends $\ttheta_i^i(t+1), \ttheta_i^j(t+1), \Lambda_{ei}^i(t), \Lambda_{ei}^j(t)$ to agent $j$ for all $j\in\mathcal{N}_i$.
\item Using values received by its neighbors, every agent $i\in\intset{n}$ updates its secondary variables for all $e=(i,j)\in E$ such that $j \in \mathcal{N}_i$:
        \begin{equation*}
            \begin{aligned}
  Z_{ei}^i(t+1) &= \frac{1}{2}\left [ 
      \frac{1}{\rho}\left(\Lambda_{ei}^i(t)+\Lambda_{ej}^i(t)\right) +
      \ttheta_i^i(t+1) + \ttheta_j^i(t+1) \right],\\
  Z_{ei}^j(t+1) &= \frac{1}{2}\left[ 
\frac{1}{\rho}\left(\Lambda_{ej}^j(t)+\Lambda_{ei}^j(t)\right) + 
      \ttheta_j^j(t+1) + \ttheta_i^j(t+1) \right].
\end{aligned}
\end{equation*}
By construction, this update maintains $Z(t+1)\in\mathcal{C}_E$.
    \item Every agent $i\in\intset{n}$ updates its dual variables for all $e=(i,j)\in E$ such that $j \in \mathcal{N}_i$:
  \begin{equation*}
    \begin{aligned}
      \Lambda_{ei}^i(t+1) &= \Lambda_{ei}^i(t) + \rho\big(\ttheta_i^i(t+1) - Z_{ei}^i(t+1)\big), \\
      \Lambda_{ei}^j(t+1) &= \Lambda_{ei}^j(t) + \rho\big(\ttheta_i^j(t+1) - Z_{ei}^j(t+1)\big).
    \end{aligned}
  \end{equation*}
\end{enumerate}

Synchronous ADMM is known to converge to an optimal solution at rate $O(1/t)$ when the objective function is convex \citep{Wei2012a}, and at a faster (linear) rate when it is strongly convex \citep{Shi2014c}. However, it requires global synchronization across the network, which can be very costly in practice.

\section{Additional Experimental Results}
\label{app:exp}

\begin{figure}[t]
    \centering
    \vskip-1em
    {\resizebox{0.5\textwidth}{!}{\input{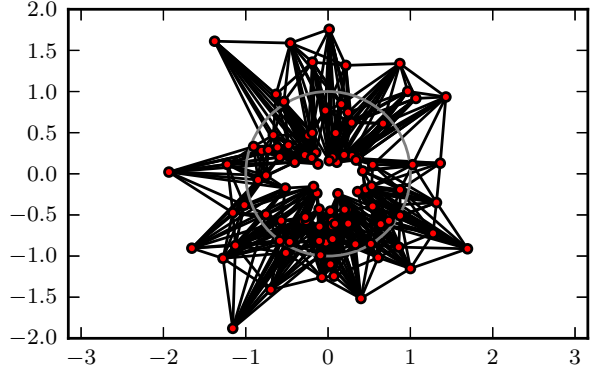}}}
      \vskip-.5em
    \caption{Target models of the agents (represented as points in $\mathbb{R}^2$) for the collaborative linear classification task. Two models are linked together when the angle between them is small, which corresponds to a small Euclidean distance after projection onto the unit circle.}
    \label{fig:cl_exp0}
\end{figure}

\paragraph{Target models in collaborative linear classification} For the experiment of Section~\ref{ssec:exp_cl}, Figure~\ref{fig:cl_exp0} shows the target models of the agents as well as the links between them. We can see that the target models can be very different from an agent to another, and that two agents are linked when there is a small enough (yet non-negligible) angle between their target models.

\begin{figure}[t]
    \centering
    {\resizebox{0.5\textwidth}{!}{\input{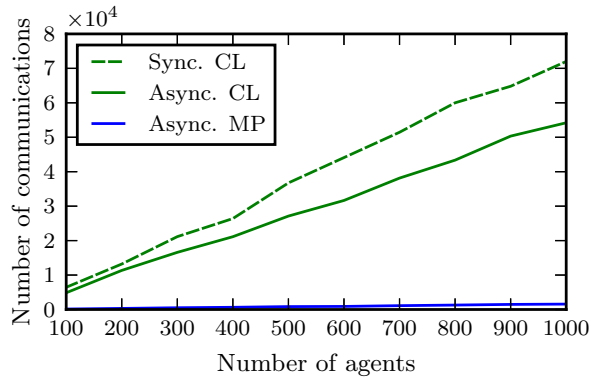}}}
      \vskip-.5em
    \caption{Number of pairwise communications needed to reach 90\% of the accuracy of the optimal models with varying number of agents (linear classification task, $p=50$).}
    \label{fig:cl_exp4}
\end{figure}

\paragraph{Scalability with respect to the number of nodes} In this experiment, we study how the number of iterations needed by our decentralized algorithms to converge to good solutions scale with the size of the network. We focus on the collaborative linear classification task introduced in Section~\ref{ssec:exp_cl} with the number $n$ of agents ranging from $100$ to $1000$. The network is a $k$-nearest neighbor graph: each agent is linked to the $k$ agents for which the angle similarity introduced in Section~\ref{ssec:exp_cl} is largest, and $W_{ij}=1$ if $i$ and $j$ are neighbors and $0$ otherwise.

Figure~\ref{fig:cl_exp4} shows the number of iterations needed by our algorithms to reach 90\% of the accuracy of the optimal set of models. We can see that the number of iterations scales linearly with $n$. In asynchronous gossip algorithms, the number of iterations that can be done in parallel also scales roughly linearly with $n$, so we can expect our algorithms to scale nicely to very large networks.

\end{document}